# Non-linear stiffness modeling of multi-link compliant serial manipulator composed of multiple tensegrity segments

*Wanda Zhao*[1], *Anatol Pashkevich*[1,2], *Damien Chablat*[1,3]

[1] *Laboratoire des Sciences du Numérique de Nantes (LS2N), UMR CNRS 6004,* Nantes, France
[2]IMT Atlantique Nantes
[3]Centre National de la Recherche Scientifique (CNRS)
Wanda.Zhao@ls2n.fr, Anatol.Pashkevich@imt-atlantique.fr, Damien.Chablat@cnrs.fr.

*Abstract*—The paper focuses on the stiffness modeling of a new type of compliant manipulator and its non-linear behavior while interacting with the environment. The manipulator under study is a serial mechanical structure composed of dual-triangle segments. The main attention is paid to the initial straight configuration which may suddenly change its shape under the loading. It was discovered that under the external loading such manipulator may have six equilibrium configurations but only two of them are stable. In the neighborhood of these configurations, the manipulator behavior was analyzed using the Virtual Joint Method (VJM). This approach allowed us to propose an analytical technique for computing a critical force causing the buckling and evaluate the manipulator shape under the loading. A relevant simulation study confirmed the validity of the developed technique and its advantages in non-linear stiffness analysis.

*Keywords – compliant manipulator; stiffness analysis; equilibrium configurations; robot buckling.*

## I. Introduction

Compliant serial manipulators have been currently used in many fields due to their flexibility, modularized construction, and low weight. Many new mechanical structures were studied in this area [1], which shown quite good performances compared with traditional rigid robots. Particular attention is attracted by tensegrity mechanisms, which are made up of a series of similar segments composed of compressive and tensile elements (cables or springs) [5], [6]. One of such structures is studied in this paper.

Some kinds of tensegrity mechanisms have been already studied carefully. In [7], the authors considered the mechanism composed of two springs and two length-changeable bars. They analyzed the mechanism stiffness using the energy method, demonstrated that the mechanism stiffness is always decreasing under external loading with the actuators locked, which may lead to "buckling" phenomenon. Also, in [8][9], the cable-driven X-shape tensegrity structures were considered; here the authors investigated the influence of cable lengths on the mechanism equilibrium configurations, which may be both stable and unstable. The relevant analysis of the equilibrium configurations as well as the stability and singularity study can be found in [10].

For robotics, the buckling phenomenon is not typical and was rarely studied before. Usually, while designing a robot, engineers prefer to avoid buckling. However, in some cases the buckling occurred but does not cause the mechanical failure. Nevertheless, such property can be also used for optimization of the mechanism architecture and useful in some other fields [12]. For this reason, for many applications, the buckling phenomenon should be obligatory taken into account in stiffness analysis. It is worth mentioning that in mechanics buckling is usually associated with the Euler column, which suddenly changes its shape when the loading force exceeds some critical value. Here, this phenomenon is studied with respect to the compliant serial structure composed of dual-triangle mechanisms.

This paper is an extension of our previous results in [11], [12][14], which concentrated on the stiffness analysis of the simplest manipulator of such type composed of two and three dual-triangle segments. It was assumed that each segment is a composition of two rigid triangle parts, which are connected by a passive joint in the center and two elastic edges on each side with controllable preload. In contrast to the previous results, here we consider a general case with an arbitrary number of segments and concentrate on the so-called "straight" initial configuration for which the external loading may cause a sudden change of the manipulator shape (i.e. buckling). For this manipulator, an analytical technique for computing a critical force causing the buckling is proposed, which was rarely mentioned before in this field. Also, the manipulator shape under the loading is analyzed in detail. The proposed approach is validated via simulation using the straightforward energy method.

## II. Mechanics of Dual-Triangle Mechanism

Let us present first a single segment of the compliant serial manipulator. It consists of two rigid triangles connected by a passive joint whose rotation is constrained by two linear springs as shown in Fig. 1, where the mechanism is symmetrical here for convenient analysis. It is assumed that the mechanism geometry is described by the triangle parameters ($a$, $b$), and the mechanism shape is defined by the central angle $q$, which is adjusted through two control inputs influencing on the springs $L_1$ and $L_2$. Let us denote the spring lengths in the non-stress state as $L_1^0$ and $L_2^0$, and the spring stiffness coefficient $k$.

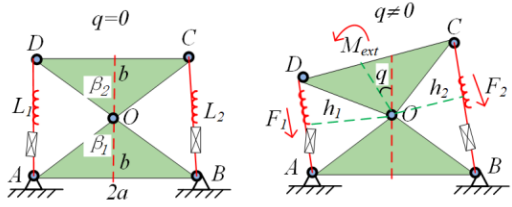

Figure 1. Geometry of a single dual-triangle mechanism.

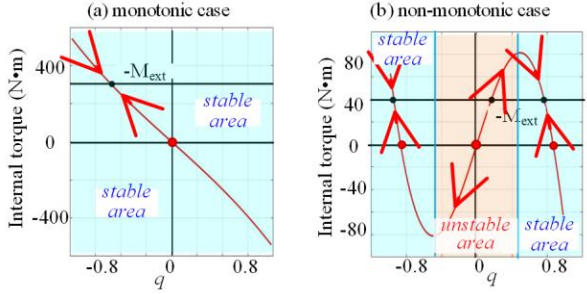

Figure 2. The torque-angle curves of the dual-triangle mechanism

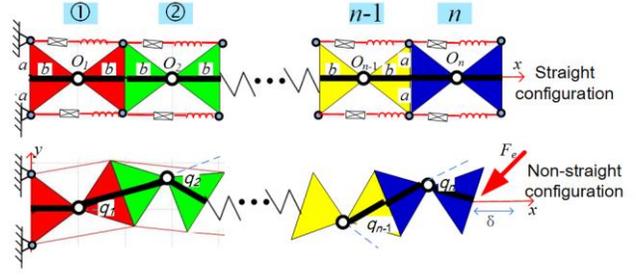

The mechanism configuration angle $q$ corresponding to the given control inputs $L_1^0$ and $L_2^0$ can be computed through the static equilibrium equation, which can be easily derived using the forces generated by the springs: $F_i = k_i(L_i - L_i^0)$, where the lengths $L_i$ are computed using the formulas $L_i = \sqrt{2c^2 + 2c^2 \cos(\theta_i)}$, where $c = \sqrt{a^2 + b^2}$ $\theta_1 = 2\beta + q$, $\theta_2 = 2\beta - q$, and $\beta = \operatorname{atan}(a/b)$. It can be proved that the corresponding torques are expressed as

$$M_1(q) = +k_1(1 - L_1^0/L_1(\theta_1))\, c^2 \sin(\theta_1)$$
$$M_2(q) = -k_2(1 - L_2^0/L_2(\theta_2))\, c^2 \sin(\theta_2)$$  (1)

Further, taking into account the external torque $M_{\text{ext}}$ applied to the moving platform, the static equilibrium equation for the considered mechanism can be written as $M(q)+M_{\text{ext}}=0$, where $M(q)= M_1(q)+ M_2(q)$ and

$$M(q) = 2ck\left[c \cos(2\beta)\sin q - L^0 \cos\beta \sin(q/2)\right] \quad (2)$$

It should be noted that the static stability of this mechanism highly depends on the equilibrium configuration defined by the angle $q$. As follows from the relevant analysis, the function $M(q)$ can be either a monotonic or non-monotonic one, so the single-segment mechanism may have multiple stable and unstable equilibriums, which are studied in detail[11], [12], [14]. As Fig. 2 shows, the torque-angle curves $M(q)$ that can be either monotonic or two-model one, the considered stability condition can be simplified and reduced to the derivative sign verification at the zero point, i.e. $M'(q)|_{q=0} < 0$, which is easy to verify in practice. So, the relevant analytical expression for the derivative

$$M'(q) = ck\left[2c \cos(2\beta)\cos q - L^0 \cos\beta \cos(q/2)\right] \quad (3)$$

allows us to present the condition of the torque-angle curve monotonicity as follows

$$L^0 > 2b \cdot \left(1 - (a/b)^2\right), \quad (4)$$

which is extensively used below.

### III. MECHANICS OF MULTI-SEGMENT MANIPULATOR

The serial manipulator considered in this paper is composed of $n$ similar sections connected in series as shown in Fig. 3, where the left-hand-side is assumed to be fixed. First, let us concentrate on the stiffness analysis of the "straight" initial unloaded configuration for which $q_i = 0, \forall i$, $(x, y) = (2nb, 0)$. This configuration is achieved by applying equal control inputs ($L_{i1}^0, L_{i2}^0, i = 1, 2,...,n$) to all mechanism segments. Under such assumptions, it is necessary to investigate the influence of the external forces $F_e = (F_x, F_y)$, which causes the end-effector displacements in the neighborhood of $(x, y) = (2nb, 0)$, moving it to a new equilibrium location $(x, y) = (2n \cdot b - \delta_x, \delta_y)$ corresponding to some nonzero configuration variables ($q_1, q_2,...,q_n$).

For this manipulator, the desired force-deflection relations $F_x(\delta_x, \delta_y)$ and $F_y(\delta_x, \delta_y)$ can be derived from the manipulator mutual kinematic and elastostatic analysis. The kinematic equations are written here as

$$
\begin{aligned}
2n \cdot b - \delta_x &= b + 2b\sum_{j=1}^{n-1}\left(\cos(\sum_{i=1}^{j} q_i)\right) + b\cos(\sum_{i=1}^{n} q_i) \\
\delta_y &= 2b\sum_{j=1}^{n-1}\left(\sin(\sum_{i=1}^{j} q_i)\right) + b\sin(\sum_{i=1}^{n} q_i)
\end{aligned} \quad (5)
$$

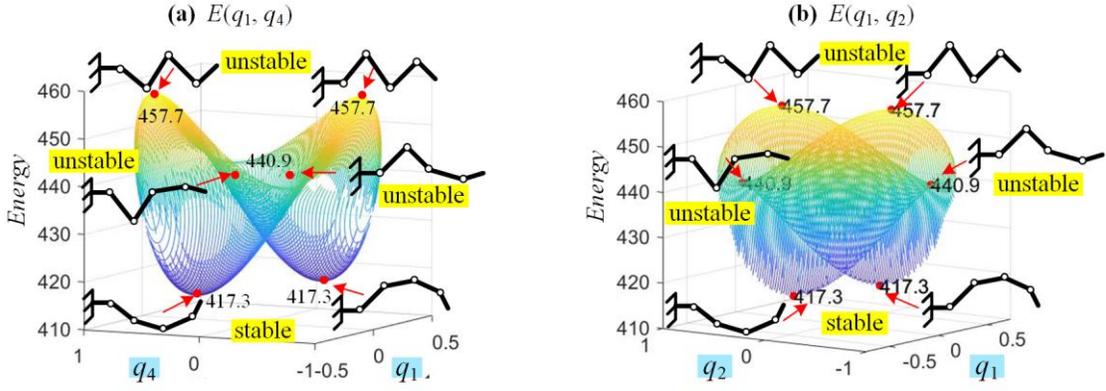

Figure 4. The energy functions $E(q_1,q_4)$, $E(q_1,q_2)$ and their critical points corresponding to the static equilibriums

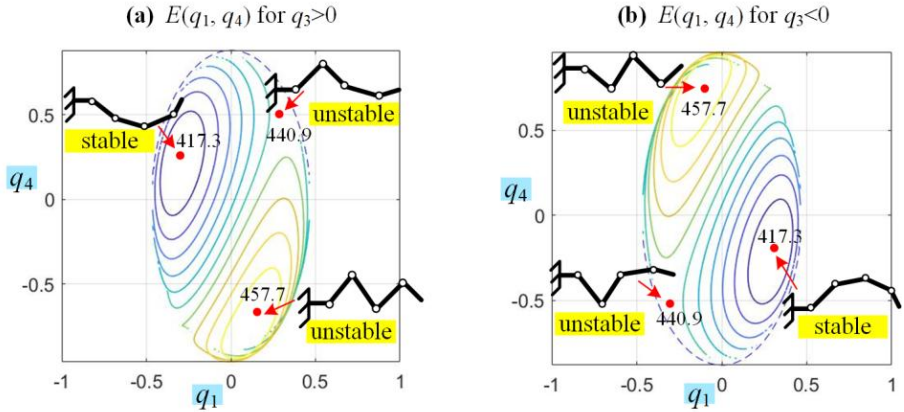

Figure 5. The contour plots of the energy functions $E(q_1,q_4)$, for different manipulator configurations

where the parameter $b$ defines the length of separate segments (see Fig. 1). It is clear that these two equations include $n$ unknown variables ($q_1, q_2, ..., q_n$), which create redundancy of the order $n$-2. This redundancy will be resolved below by applying the minimum elastic energy principle allowing to find the desired equilibrium configuration angles using the following matrix equation

$$\begin{bmatrix} M_{q1} \\ ... \\ M_{qn} \end{bmatrix} + \begin{bmatrix} \mathbf{J_q^T} \end{bmatrix}_{n\times 2} \cdot \begin{bmatrix} F_x \\ F_y \end{bmatrix} = \begin{bmatrix} 0 \\ ... \\ 0 \end{bmatrix} \quad (6)$$

which includes the manipulator Jacobian

$$\mathbf{J_q} = b \cdot \begin{bmatrix} -\sum_{j=1}^{N}\left(\eta_j \sin \sum_{i=1}^{j} q_i\right) & ... & -\sum_{j=N}^{N}\left(\eta_j \sin \sum_{i=1}^{j} q_i\right) \\ \sum_{j=1}^{N}\left(\eta_j \cos \sum_{i=1}^{j} q_i\right) & ... & \sum_{j=N}^{N}\left(\eta_j \cos \sum_{i=1}^{j} q_i\right) \end{bmatrix}_{2\times n} \quad (7)$$

where $\eta_j = 2$ for $j<n$ and $\eta_j =1$ for $j=n$. So totally, combining both geometric and elastostatic equations (5) and (6) one can obtain $n$+2 nonlinear equations for $n$+2 unknowns $q_i, \forall i$ and $F_x, F_y$, assuming that $(\delta_x, \delta_y)$ are known.

Obviously, in the general cases such a nonlinear system can only be solved numerically, using Newton's method for example. However, for relatively small $n$ it is possible to apply the semi-analytical technique. For instance, if $n=4$ the geometric model (5) allows us to reduce analytically the number of unknown variables down to two. In particular, if the angles $q_1$ and $q_2$ are assumed to be known, the remaining ones $q_3$ and $q_4$ can be computed from the classical inverse kinematics of the two-link manipulator. Further, it is possible to compute the two-variable energy function $E(q_i,q_j)$ and find its minimums and maximums numerically, which define the stable and unstable equilibriums respectively. Examples of such computations are presented in Figs. 4 and 5, where the independent variables are $(q_1,q_4)$ and $(q_1,q_2)$.

IV. MANIPULATOR STIFFNESS BEHAVIOR

As follows from the energy plots presented in Fig. 4, in each case there are two global maximums and two global minimums corresponding to the stable and unstable equilibriums respectively. Besides, there are also two saddle points here that were discovered after numerical analysis of the energy function gradient. These saddle points also correspond to the unstable equilibriums. So in this case study, the energy functions $E(q_i, q_j)$ allow detecting six equilibriums: two stable ones with "U-shape" of the manipulator, two unstable with "Z-shape" and two unstable with "ZU-shape". However, it is worth mentioning that for other combinations of the geometric parameters ($a/b$, $L^o/b$) and end-effector deflections ($\delta x$, $\delta y$), the energy function may have a higher number of critical points.

This approach allows us also to obtain the force-deflection relations. In fact, the energy minimums defining stable equilibriums provide us with the configuration angles $(q_1, q_2, ..., q_n)$ allowing to find the force $(F_x, F_y)$ corresponding to the given displacement $(\delta_x, \delta_y)$. From the expression

$$\begin{bmatrix} F_x \\ F_y \end{bmatrix} = -\begin{bmatrix} \mathbf{J_q^T} \end{bmatrix} \cdot \begin{bmatrix} \mathbf{J_q J_q^T} \end{bmatrix}^{-1} \cdot \begin{bmatrix} M_{q1} \\ ... \\ M_{qn} \end{bmatrix} \quad (8)$$

which is based on the Moore-Penrose pseudo-inverse of the manipulator Jacobian $\mathbf{J_q}$ and the joint torques expressions $M_{qi}$ from (2) for the configuration angles of the stable equilibriums. Applying such techniques for different $(\delta_x, \delta_y)$ one can get the desired force-deflection relations $F_x(\delta_x, \delta_y)$ and $F_y(\delta_x, \delta_y)$.

Examples of such curves for the case $n=4$ and ($\delta x$=var, $\delta y$=0) are presented in Fig. 6, which clearly demonstrate the jump discontinuity at the beginning of the curve $F_x(\delta_x, 0)$. Hence, the stiffness properties of the considered multi-segment manipulator are essentially nonlinear and force-deflection relations are discontinuous, which is observed physically as the "buckling" phenomenon.

To compute the critical force causing the buckling, let us apply the following analytical technique. First, let us linearize the geometric model (5)

$$\begin{aligned} \delta_x/b &= \sum_{j=1}^{n-1}\left(\sum_{i=1}^{j} q_i\right)^2 + \frac{1}{2}\left(\sum_{i=1}^{n} q_i\right)^2 \\ \delta_y/b &= 2\sum_{j=1}^{n-1}\left(\sum_{i=1}^{j} q_i\right) + \sum_{i=1}^{n} q_i \end{aligned} \quad (9)$$

and to present the Jacobian as

$$\mathbf{J_q} = b\begin{bmatrix} -\sum_{j=1}^{n}\left(\eta_j \sum_{i=1}^{j} q_i\right) & -\sum_{j=2}^{n}\left(\eta_j \sum_{i=1}^{j} q_i\right) & ... & -\sum_{j=n}^{n}\left(\eta_j \sum_{i=1}^{j} q_i\right) \\ 2n-1 & 2n-3 & ... & 1 \end{bmatrix}_{2\times n} \quad (10)$$

that can be also expressed in matrix form as follows

$$\mathbf{J_q} \approx b \cdot [\mathbf{S_1} \cdot \mathbf{q} \mid \mathbf{S_0}]^T \quad (11)$$

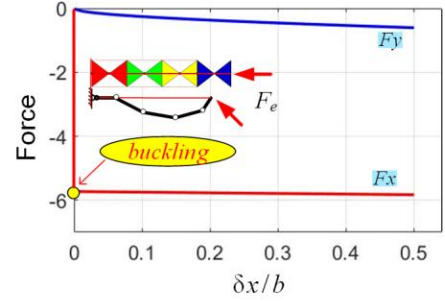

Figure 6. The force-deflection curves for the 4-link manipulator with the geometric parameters $a/b$=1.0, $L^0/b$=1.0, $k=1$, ($\delta x$=var, $\delta y$=0).

For example, for $n=4$ the above Jacobian and the matrices $\mathbf{S_1}$, $\mathbf{S_0}$ are written as

$$\mathbf{J_q^T} = b \cdot \begin{bmatrix} -7q_1-5q_2-3q_3-q_4 & 7 \\ -5q_1-5q_2-3q_3-q_4 & 5 \\ -3q_1-3q_2-3q_3-q_4 & 3 \\ -q_1-q_2-q_3-q_4 & 1 \end{bmatrix}; \quad \mathbf{S_1} = -\begin{bmatrix} 7 & 5 & 3 & 1 \\ 5 & 5 & 3 & 1 \\ 3 & 3 & 3 & 1 \\ 1 & 1 & 1 & 1 \end{bmatrix} \quad (12)$$

and $\mathbf{S_0} = \begin{bmatrix} 7 & 5 & 3 & 1 \end{bmatrix}^T$. The latter allows us to rewrite the static equilibrium equation (6) as

$$K_{eq} b^{-1} \cdot \mathbf{q} + \mathbf{S_1} \mathbf{q} \cdot F_x + \mathbf{S_0} \cdot F_y = \mathbf{0} \quad (13)$$

where $K_{eq} = dM_{qi}/dq_i$, $\forall i$ is the equivalent stiffness of the manipulator joints computed by linearization $M_{qi} \approx K_{eqi} \cdot q_i$ of the torque expression (2) for $q_i \to 0$ which yields

$$K_{eq} = \frac{k}{b}\left[2\left(b^2-a^2\right)-bL^0\right] \quad (14)$$

Then, let us assume that $\delta y$=0 and combine corresponding constraints obtained from the geometric model (9)

$$\begin{bmatrix} 2n-1 & 2n-3 & ... & 1 \end{bmatrix} \cdot \mathbf{q} \equiv \mathbf{S_0^T q} = 0 \quad (15)$$

with the static equilibrium condition (6), which yields the matrix equation with $n+2$ unknowns $q_1, ..., q_n$ and $F_x, F_y$

$$\begin{bmatrix} \mathbf{S_1} & \mathbf{0}_{n\times 1} \\ \mathbf{0}_{1\times n} & 0 \end{bmatrix} \cdot \begin{bmatrix} \mathbf{q} \\ F_y \end{bmatrix} \cdot F_x + \begin{bmatrix} K_{eq} b^{-1} \mathbf{I}_{n\times n} & \mathbf{S_0} \\ \mathbf{S_0^T} & 0 \end{bmatrix} \cdot \begin{bmatrix} \mathbf{q} \\ F_y \end{bmatrix} = \mathbf{0} \quad (16)$$

that can be further rewritten in the form

$$(\mathbf{A} \cdot F_x + \mathbf{B})\mathbf{v} = \mathbf{0} \quad (17)$$

where

$$\mathbf{A} = \begin{bmatrix} \mathbf{S_1} & \mathbf{0}_{n\times 1} \\ \mathbf{0}_{1\times n} & 0 \end{bmatrix}; \quad \mathbf{B} = \begin{bmatrix} K_{eq} b^{-1} \mathbf{I}_{n\times n} & \mathbf{S_0} \\ \mathbf{S_0^T} & 0 \end{bmatrix}; \quad \mathbf{v} = \begin{bmatrix} \mathbf{q} \\ F_y \end{bmatrix} \quad (18)$$

Table. 1 Nonzero eigenvalues and corresponding eigenvectors of matrix $\mathbf{B}^{-1}\mathbf{A}$
for 4-segment manipulator with the geometric parameters $a/b=1.0$, $L^0/b=1.0$, $k=1$.

|   | $\lambda$ | $\alpha_1$ | $\alpha_2$ | $\alpha_3$ | $\alpha_4$ | $\alpha_5$ |
|---|---|---|---|---|---|---|
| #1 | −1.746 | 0.525 | -0.227 | -0.719 | -0.388 | -0.075 |
| #2 | −0.734 | 0.352 | -0.707 | 0.162 | 0.590 | -0.050 |
| #3 | −0.520 | 0.124 | -0.387 | 0.589 | -0.699 | -0.018 |

It can be easily seen that, the obtained matrix equation (17) with unknowns $\mathbf{v} \in \mathbf{R}^{n+1}$ and $F_x \in \mathbf{R}^1$ is similar to the equation considered in the classical matrix analysis for computing the matrix eigenvectors and eigenvalues. In fact, it can be proved that the matrix $\mathbf{B}$ is invertible, so (17) can be presented in the standard form as

$$(\mathbf{B}^{-1}\mathbf{A} - \lambda \cdot \mathbf{I})\mathbf{v} = \mathbf{0}; \qquad \lambda = -1/F_x \qquad (19)$$

Hence, the desired critical force $F_x^0$ causing the buckling can be computed using the largest (in absolute value) eigenvalue of the matrix $\mathbf{B}^{-1}\mathbf{A}$

$$F_x^0 = -\left(\max_i |\lambda_i|\right)^{-1} \qquad (20)$$

which corresponds to the smallest amplitude of the external force $F_x$ ensuring the manipulator equilibrium with $\mathbf{q} \neq \mathbf{0}$. For example, for $n=4$ and $K_{eq}b^{-1}=1$, relevant computing yields the following eigenvalues

$$\lambda_i \in \{-1.746, -0.734, -0.520, 0, 0\}$$

whose eigenvectors are presented in Table 1. It can be also proved that for $n \geq 3$, there are exactly $n-1$ nonzero eigenvalues here.

Using the obtained eigenvectors $\mathbf{v}_1,...,\mathbf{v}_{n-1}$ corresponding to the nonzero eigenvalues, it is possible to express in parametric form the variables $q_i$ and $F_y$ as

$$q_i = \alpha_i \cdot t, \ (i=1,...,n); \qquad F_y = \alpha_{n+1} \cdot t \qquad (21)$$

where $\alpha_i$ are the components of the eigenvector $\mathbf{v} = [\alpha_1,...,\alpha_{n+1}]^T$ and $t$ is some small number. This presentation allows us to express the manipulator elastostatic energy in equilibrium configuration as

$$E_{eq} = \frac{1}{2}K_{eq}\sum_{i=1}^{n}\alpha_i^2 \cdot t^2 \qquad (22)$$

and also to compute the corresponding deflection $\delta x$

$$\delta_x/b = \sum_{j=1}^{n-1}\left(\sum_{i=1}^{j}\alpha_i\right)^2 t^2 + \frac{1}{2}\left(\sum_{i=1}^{n}\alpha_i\right)^2 t^2 \ @ \ \mu_x \cdot t^2 \qquad (23)$$

The latter allows us to compare the elastostatic energy corresponding to different equilibriums with the same $\delta_x$, which defines the parameters $t = \sqrt{\delta_x b^{-1} \mu_x^{-1}}$ and leads to the following expression for the energy

$$E_{eq} = \mu_{eq} \cdot K_{eq} \cdot \delta_x / 2b \qquad (24)$$

where

$$\mu_{eq} = \sum_{i=1}^{n}\alpha_i^2 \Bigg/ \left(\sum_{j=1}^{n-1}\left(\sum_{i=1}^{j}\alpha_i\right)^2 + \frac{1}{2}\left(\sum_{i=1}^{n}\alpha_i\right)^2\right) \qquad (25)$$

Besides, such presentation allows us to express the forces $F_x$, $F_y$ in the equilibrium neighborhood after buckling as

$$F_x \approx F_x^0; \quad F_y \approx \frac{\alpha_{n+1}}{\sqrt{b\mu_x}} \cdot \sqrt{\delta_x} \qquad (26)$$

which is in good agreement with Fig. 6, where the curve $F_x(\delta_x)$ is quasi-linear and the shape of the curve $F_y(\delta_x)$ follows to the shape of $\sqrt{\delta_x}$.

Table 2 Possible manipulator shapes in static equilibrium after the buckling for $n=4$.

| | q1 | q2 | q3 | q4 | Geometric shape | | Stability | Energy factor $\mu_{eq}$ |
|---|---|---|---|---|---|---|---|---|
| Case #1 q1<0 | − | + | + | + | U shape: | 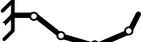 | stable | 1.1447 |
| | − | + | − | + | Z shape: | 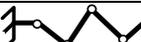 | unstable | 3.8429 |
| | − | + | − | − | ZU shape: | 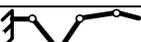 | unstable | 2.7272 |
| Case #2 q1>0 | + | − | − | − | U shape: | 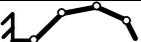 | stable | 1.1447 |
| | + | − | + | − | Z shape: | 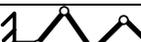 | unstable | 3.8429 |
| | + | − | + | + | ZU shape: | 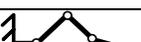 | unstable | 2.7272 |

The parameterized presentation of the joint angles $q_i = \alpha_i \cdot t$ allows us also to evaluate the manipulator shape in the possible equilibrium configurations. In fact, the matrix $\mathbf{B}^{-1}\mathbf{A}$ provides us with $n-1$ different sets of $\{\alpha_1,...,\alpha_n\}$ corresponding to nonzero eigenvalues. Each of such set yields to two symmetrical equilibriums (for $t>0$ and $t<0$), whose shape can be evaluated by analyzing the signs of $\alpha_i$. Hence, the total number of the different equilibriums is equal to $2(n-1)$, but two of them providing the minimum of the elastostatic energy are globally stable. Examples of possible manipulator shapes in static equilibrium after the buckling for $n$=4 are presented in Table 2, where the "U-shape" is stable and "Z-shape" is unstable. Besides, here there is an additional "ZU-shape" that is also unstable.

## V. CONCLUSION

The paper focuses on the stiffness analysis of a new type of compliant serial manipulator composed of dual-triangle segments, which is a specific case of the tensegrity mechanisms that currently are widely used in soft robotics. The main attention is paid to the initial straight configuration of the manipulator, which may suddenly change its shape under the loading. A similar property is known from the Euler column mechanics where it is called buckling. It was proved that under the external loading such manipulator may have six equilibrium configurations but only two of them are stable. To find these equilibriums, both the straightforward energy method and the proposed analytical technique were used. The latter of them is based on the VJM approach and allowed to find the relations between the end-effector deflection and the external force, as well as to compute a critical force causing the buckling analytically. It should also mentioned that such method can be generally used for the similar structure serial manipulators. Besides, the manipulator shape under the loading was analyzed in detail. Relevant simulation study and comparison with the energy method confirmed the validity of the developed technique and demonstrated its advantages in non-linear stiffness analysis. In the future, this technique will be used for development of relevant control algorisms and related redundancy resolution.

ACKNOWLEDGMENT

This work was supported by the China Scholarship Council ( No. 201801810036 ).